%% file: conference_101719.tex
\documentclass[a4paper]{article}

\input{1_preamble}

\begin{document}

\title{Spatiotemporal Graph Neural Networks in short-term load forecasting: Does adding Graph Structure in Consumption Data Improve Predictions?}

\author{
Quoc Viet NGUYEN,
Joaquin DELGADO FERNANDEZ, \\
Sergio POTENCIANO MENCI \\
\textit{Interdisciplinary Centre for Security, Reliability and Trust - SnT,} \\
\textit{University of Luxembourg, Luxembourg }\\
\{quocviet.nguyen, 
joaquin.delgadofernandez,
sergio.potenciano-menci\}@uni.lu
\thanks{This research was funded in part by the Luxembourg National Research Fund (FNR) and PayPal, PEARL grant reference 13342933/Gilbert Fridgen and by FNR grant reference HPC BRIDGES/2022\_Phase2/17886330/DELPHI. For the purpose of open access and in fulfillment of pen access and fulfilling the obligations arising from the grant agreement, the author has applied a Creative Commons Attribution 4.0 International (CC BY 4.0) license to any Author Accepted Manuscript version arising from this submission.
The experiments presented in this paper were carried out using the HPC
facilities of the University of Luxembourg~\cite{ORBi-aab35225-a6bc-496d-a6e7-621189ebff46}– see hpc.uni.lu
}
}

\maketitle

\input{sections/00__abstract}

\input{sections/01__introduction}

\input{sections/02__background}

\input{sections/03__experiments}

\input{sections/04__results}

\input{sections/05__conclusions}

\section*{Declaration of Generative AI and AI-assisted technologies in the writing process}
Statement: During the preparation of this work, the author(s) used ChatGPT~\cite{ChatGPT} to paraphrase and fix grammatical mistakes. After using this tool/service, the author(s) reviewed and edited the content as needed and take(s) full responsibility for the content of the publication.

\bibliographystyle{IEEEtran}
\bibliography{reference}

\end{document}

%% file: 1_preamble.tex
\usepackage{amsmath,amssymb,amsfonts}
\usepackage{algorithmic}
\usepackage{graphicx}
\usepackage{textcomp}
\usepackage{multirow}
\usepackage[a4paper, total={6in, 8in}]{geometry}
\usepackage[dvipsnames]{xcolor}
\usepackage{tabularx}
\usepackage{url} 
\usepackage[hidelinks]{hyperref}
\usepackage{dblfloatfix}
\def\BibTeX{{\rm B\kern-.05em{\sc i\kern-.025em b}\kern-.08em
    T\kern-.1667em\lower.7ex\hbox{E}\kern-.125emX}}
    
\usepackage[acronym, nonumberlist]{glossaries}
\usepackage{adjustbox}
\usepackage[most]{tcolorbox}

\newtcbox{\inlinebox}[1][]{
    colframe=Cyan,    
    colback=white,       
    arc=0.5mm,             
    boxrule=0.8pt,       
    width=0.3cm,
    height=0.6cm,        
    valign=center,       
    nobeforeafter,       
    #1                   
}

\providecommand{\keywords}[1]
{
  \small	
  \textbf{\textit{Keywords---}} #1
}

\input{glossary}



%% file: glossary.tex
\newacronym{stlf}{STLF}{Short-term Load Forecasting}

\newacronym[plural=STGNNs, firstplural=Spatiotemporal Graph Neural Networks (STGNNs)]{stgnn}{STGNN}{Spatiotemporal Graph Neural Network}

\newacronym{tts}{TTS}{Time-then-Space}

\newacronym{tas}{T\&S}{Time-and-Space}

\newacronym{rnn}{RNN}{Recurrent Neural Network}

\newacronym{gru}{GRU}{Gated Recurrent Unit}

\newacronym{cnn}{CNN}{Convolutional Neural Network}

\newacronym{ms}{MS}{Message Passing}

\newacronym{gnn}{GNN}{Graph Neural Network}

\newacronym{var}{VAR}{Vector Autoregressive}

\newacronym{stegnn}{STEGNN}{ Spatial-Temporal Embedding Graph Neural Network}

\newacronym{gcn}{GCN}{Graph Convolutional Network}

\newacronym{lstm}{LSTM}{Long Short-Term Memory}

\newacronym{stt}{STT}{Space-then-Time}

\newacronym{mlp}{MLP}{Multilayer Perceptron}

\newacronym{lcl}{LCL}{Low Carbon London}

\newacronym{mae}{MAE}{Mean Absolute Error}

\newacronym{mape}{MAPE}{Mean Absolute Percentage Error}

\newacronym{rmse}{RMSE}{Root Mean Square Error}

\newacronym{tfm}{Transformer}{Transformer}

\newacronym{grugcn}{GRUGCN}{GRUGCN}

\newacronym{gcgru}{GCGRU}{GCGRU}

\newacronym{t-gcn}{T-GCN}{T-GCN}

\newacronym{agcrn}{AGCRN}{AGCRN}

\newacronym{gw}{GraphWavenet}{GraphWavenet}

\newacronym{fc-gnn}{FC-GNN}{Fully-Connected Graph Neural Network}

\newacronym{bp-gnn}{BP-GNN}{Bipartite-Graph Neural Network}

%% file: sections/00__abstract.tex
\begin{abstract}
\acrfull{stlf} plays an important role in traditional and modern power systems. 
Most \acrshort{stlf} models predominantly exploit temporal dependencies from historical data to predict future consumption. Nowadays, with the widespread deployment of smart meters, their data can contain spatial-temporal dependencies. In particular, their consumption data is not only correlated to historical values but also to the values of 'neighboring' smart meters. This new characteristic motivates researchers to explore and experiment with new models that can effectively integrate spatial-temporal interrelations to increase forecasting performance. \glspl{stgnn} can leverage such interrelations by modeling relationships between smart meters as a graph and using these relationships as additional features to predict future energy consumption.
While extensively studied in other spatiotemporal forecasting domains such as traffic, environments, or renewable energy generation, their application to load forecasting remains relatively unexplored, particularly in scenarios where the graph structure is not inherently available. This paper overviews the current literature focusing on \glspl{stgnn} with application in \acrshort{stlf}.
Additionally, from a technical perspective, it also benchmarks selected \acrshort{stgnn} models for \acrshort{stlf} at the residential and aggregate levels. The results indicate that incorporating graph features can improve forecasting accuracy at the residential level; however, this effect is not reflected at the aggregate level. 
\end{abstract}

\keywords{Benchmark, \acrlong{stlf}, graph neural network, spatiotemporal forecasting.}

%% file: sections/01__introduction.tex
\section{Introduction}\label{sec:intro}
 \acrfull{stlf} plays a vital role in power systems by supporting grid and market operations and, consequently, helping with the reliability of traditional and modern power systems~\cite{eren2024comprehensive}. Its importance grows as power systems become more complex and decentralized~\cite{big_data_smart_grid}. In particular, modern power systems require better forecasts to deal with the increasing grid (e.g. dispatch, reconfiguration) and market operations (e.g. portfolio balancing) caused by dynamic fluctuations in generation and consumption~\cite{big_data_smart_grid}. Tackling the dynamic fluctuations appropriately beforehand also has a significant economic impact affecting all power system players. For instance, a one-percent increase in forecast accuracy could save up to £10 million annually in the UK~\cite{arastehfar_short-term_2022}.

In general, most \acrshort{stlf} models emerge from statistical or machine learning models~\cite{hong_probabilistic_2016}. Statistical models usually assume the property of time series such as stationarity or invertibility~\cite{arastehfar_short-term_2022}, which might be unsuitable to model volatile energy consumption data down at the residential level~\cite{Moustati2024-oo}. The alternative is to shift to data-driven approaches, such as machine learning and especially deep learning, to effectively model the expanding space of available data, mainly caused by the introduction of smart meters.  
Currently, most \acrshort{stlf} models rely solely on the temporal dependency of historical data for forecasting. However, since consumption data can be collected from multiple smart meters at the household level, the interrelation between different households can also be discovered and used~\cite{WU2023125939}. Incorporating information from 'neighboring' households with strong interconnections could potentially improve forecasting accuracy for a specific household. Consequently, forecasting models should integrate both temporal and spatial dependencies to effectively capture the data's dynamics.

A solution to capture these dependencies is to use \acrfull{stgnn} because, in addition to processing data in temporal order, \acrshort{stgnn} accounts for spatial dependencies through a graph-based approach. 
This deep learning-based method has been the subject of substantial research in fields such as traffic prediction, environmental studies, and energy generation, where spatio-temporal characteristics are evident~\cite{bui_spatial-temporal_2022}. 
Recently, more energy research has applied this architecture to forecast energy consumption at the residential level~\cite{arastehfar_short-term_2022, lin_residential_2021}. Although graphs in other problems such as traffic or energy generation can be derived from geographic locations, spatial proximity does not fully reflect the similarity in energy consumption patterns. This is because consumption behaviors at the residential level are stochastic, and the similarity between usage patterns lies on sociodemographic factors rather than spatial proximity~\cite{Feng2023STGNetSR}. To represent the relationship between households, many \acrshort{stlf} studies have directly extracted the similarity of signals among them~\cite{bloemheuvel_graph_2024}. Another, more flexible way is to model graph structure as learnable parameters and optimize it during training to produce the best forecast~\cite{lin_residential_2021,wei_short-term_2023}. These strategies add the spatial notion in the energy consumption data as a graph and make the application of \glspl{stgnn} on the \acrshort{stlf} problem feasible. However, the absence of an inherent graph structure in energy consumption behaviors necessitates constructing one from historical data. This raises an important question: \textit{Does a temporally informed graph model predict more effectively load that a model based solely on temporal features?}

This question combined with increasing research on \acrshort{stgnn} for \acrshort{stlf} has motivated our research. Our goal is to summarize the current literature on \acrshort{stlf} using \glspl{stgnn} and to identify the key components that influence the performance of the models by:
\begin{itemize}
    \item Evaluating the performance of representative \acrshort{stgnn} models in residential \acrshort{stlf}. 
    \item Identifying relevant factors in the construction of \glspl{stgnn} that affect performance in \acrshort{stlf}. 
\end{itemize}

The remainder of the paper is structured as follows. Section \ref{sec:background} gives a brief overview of the existing \acrshort{stgnn} architectures and models for spatiotemporal forecasting in general and load forecasting in particular. In Section \ref{sec:exp_result}, we designed experiments to validate and compare the performance of \acrshort{stgnn} on \acrshort{stlf} in different time scales. Building upon these results, we provide insight and explanations of the results specific to load forecasting. The study concludes with a summary and future directions in Section \ref{sec:conclusion}.

%% file: sections/02__background.tex
\section{Overview of STGNN for STLF}\label{sec:background}


\glspl{stgnn} are models designed to handle time series data collected from various locations~\cite{cini_graph_2023}. In the context of \acrshort{stlf}, we consider a dataset collected from $N$ consumers (e.g. households). In the simplest case, we assume 
for each consumer, data only contain historical energy consumption from smart meters. 
Let $x^i_t \in \mathbb{R}$ be the energy consumption of consumer $i$ at time step $t$; each time series $\{x^i_t\}_{t:t+T}$ is the energy consumption of consumer $i$ in period $t \rightarrow t+T$. 
Consequently, by stacking all consumers, the matrix $\mathbf{X}_{t:t+T} \in \mathbb{R}^{N \times T}$ represents the consumption records of N consumers in the period $t \rightarrow t+T$. Given the consumption data $\mathbf{X}_{t-W:t}$ from $W$ previous steps , \acrshort{stgnn} models forecast consumption $\hat{\mathbf{X}}_{t:t+H}$ in $H$ next steps for all consumers. 

In doing so, \acrshort{stgnn} represents consumers and their relationships by a graph structure. A graph $\mathcal{G} = (\mathcal{V}, \mathcal{E})$ consists of a set of nodes $\mathcal{V} = \{v_1,v_2,...,v_N\}$ and a set of edges $\mathcal{E} \subseteq \mathcal{V} \times \mathcal{V}$, where $(v_i,v_j) \in \mathcal{E}$ if node $v_i$ connects to node $v_j$. This connectivity is compactly represented by an adjacency matrix $\mathcal{A}$, where an entry $a_{ij} > 0$ signifies the edge weight between nodes $v_i$ and $v_j$. In the context of residential \acrshort{stlf}, each household is assigned to one node and its features contain time series  of its historical energy consumption. Respectively, the edges can be derived based on the patterns between residential load profiles~\cite{wang_short-term_2023}. 

In what follows, we first describe how graph structures are usually constructed, then detail the components of \acrshort{stgnn} and how those components interact with each other. Finally, we present some representative models in the literature.

\subsection{Graph Formation}\label{subsec:graph}
To capture the spatial dependency between different nodes, it is necessary to provide a spatial structure in the form of a graph. The topology of the graph dictates how the features are aggregated between the nodes. Based on how the graph is constructed, the graph formation methods in the literature can be classified as follows~\cite{bloemheuvel_graph_2024}:
\begin{itemize}
    \item \textit{Predefined graph}: The topology of the graph is fixed during training. The edge of the graph may be established using supplementary information and by assessing the similarity between time series. The similarity measure can be based on Pearson coefficient~\cite{fernandez_privacy-preserving_2022}, Euclidean distance, DTW distance~\cite{10202782}, or correntropy~\cite{cini_graph-based_2023}. An edge is considered to exist when the similarity surpasses a defined threshold value~\cite{bloemheuvel_graph_2024}.
    \item \textit{Learnable graph}: Some models integrate graph formation into the learning process. This technique does not require a graph from the dataset before training but self-organizes the graph during training so that it can facilitate the flow of information for graph neural network~\cite{wu_graph_nodate}. We identify some learning algorithms that incorporate graph structure learning for downstream tasks (i.e., forecasting) later in section \ref{subsec:exampleSTGNN}.
\end{itemize}

\subsection{Temporal and Spatial Processing Unit as components of STGNN}\label{subsec:processing}

To model spatiotemporal data as in load forecasting, one must process information in temporal and spatial dimensions. 
The most popular deep learning models to process temporal information are through \acrfull{rnn}~\cite{sutskever_sequence_2014}, \acrfull{cnn} models, or \acrfull{mlp}~\cite{rodrigues_short-term_2023}. 

Respectively, the most dominant method to propagate information along the graph is through the \acrfull{ms} paradigm~\cite{cini_graph_2023} which involves 2 steps:
\begin{enumerate}
    \item \textit{Message Aggregation}: Each node collects information (or "messages") from its neighboring nodes. This step captures the local structure and features of the graph by pooling or combining the attributes of connected nodes. 
    \item \textit{Feature Update}: After aggregation, each node updates its own representation by combining the aggregated information with its existing attributes. This step refines the node's state, embedding its local and neighboring information into a new feature representation.
\end{enumerate}

There are several models fit into this paradigm, but one of the most popular is \acrfull{gcn}~\cite{mansoor_spatio-temporal_2023}. It is present in numerous examples of \acrshort{stgnn} within our study.

\subsection{STGNN architecture}\label{subsec:architecture}
We refer to the structure of \glspl{stgnn} in the literature. The most prominent architectures of STGNN are \acrfull{tts}, \acrfull{tas}~\cite{gao_equivalence_2022}.

\subsubsection{Time-then-Space (TTS) architecture}\label{subsubsec:tts}
In \acrshort{tts} architecture, the models encode information in the temporal dimension first, as depicted by the solid arrow in Fig. \ref{fig:tts}. The abstract representation of each node is now broadcast (dashed arrow in Fig. \ref{fig:tts}) to their neighbors through the spatial unit to incorporate useful information in the spatial dimension.

\subsubsection{Time-and-Space (T\&S) architecture}\label{subsubsec:tas}
\begin{figure}[ht]
    \centering
    \begin{minipage}[t]{0.46\textwidth} 
        \centering
        \includegraphics[width=\linewidth]{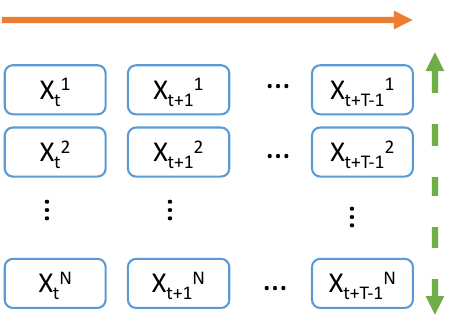} 
        \caption{\acrshort{tts} architecture.}
        \label{fig:tts}
    \end{minipage}
    \hfill 
    \begin{minipage}[t]{0.48\textwidth} 
        \centering
        \includegraphics[width=\linewidth]{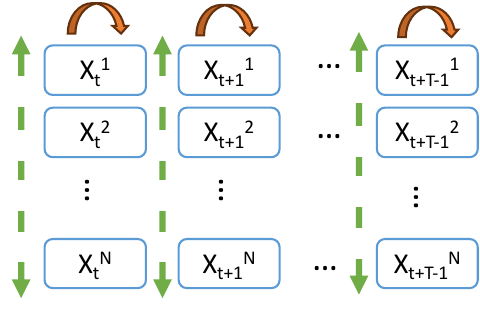} 
        \caption{\acrshort{tas} architecture.}
        \label{fig:tas}
    \end{minipage}
    \label{fig:architectures}
\end{figure}

Alternatively, the \acrshort{tas} architecture integrates the processing of temporal and spatial features more cohesively. At each time step, the features of individual nodes are propagated and abstracted through a spatial processing unit, updating the node representations in the spatial dimension (dashed arrow in Fig. \ref{fig:tas}). Then, the node representation is processed using a recurrent unit, producing a hidden state at that time step.  
Subsequently, this internal hidden state is combined with the upcoming observation to produce the hidden state in the next step (as illustrated by the curved arrow in Fig. \ref{fig:tas}). It is worth noting that a recurrent model is employed as the temporal processing unit in this context~\cite{gao_equivalence_2022}.



\subsection{Examples of STGNN models}\label{subsec:exampleSTGNN}
Given the different architectures of \acrshort{stgnn} models, we review  existing models that are representatives of the 
 described architectures. We present and compare the similarity between models based on the components and architectures in Section \ref{subsec:processing} and \ref{subsec:architecture}.

\paragraph{\acrshort{grugcn}~\cite{gao_equivalence_2022}} This model is of type \acrshort{tts}. It uses the \acrfull{gru} (a variant of \acrshort{rnn}) as a temporal processing unit to represent temporal characteristics of each node and then applies \acrshort{gcn} on top of the encoded features to account for spatial dependencies.

\paragraph{\acrshort{gcgru}~\cite{arastehfar_short-term_2022}} This model uses \acrshort{tas} architecture. In particular, it uses \acrshort{gcn} as the spatial processing unit to capture the spatial dependency at time step $t$. This unit acts as a cell in the \acrshort{gru} model, which recursively calculates the hidden state of the entire graph. Note that for a more consistent comparison, we replace the \acrfull{lstm} model in \cite{arastehfar_short-term_2022} by \acrshort{gru} as in other models.

\paragraph{\acrshort{t-gcn}~\cite{huang_gated_2023}} This model updates node features by a 2-layer GCN before processing them by \acrshort{gru} model. It is similar to \acrshort{gcgru} but updates only node features through the \acrshort{gcn}, not the hidden state.

\paragraph{\acrshort{agcrn}~\cite{bai_adaptive_2020}} This model is similar to \acrshort{gcgru} in terms of encoding both spatial and temporal dimensions. However, to model the relationships between nodes more flexibly, this approach introduces a learnable embedding for each node. Consequently, the feature in each node is not only determined by the historical data but also by the embedding. This allows the model to determine the optimal embeddings that maximize the forecast performance.

\paragraph{\acrshort{gw}~\cite{lin_spatial-temporal_2021}} Similar to \acrshort{agcrn}, this model assigns each node a learnable embedding. In terms of forecasting, it employs a temporal convolutional layer to encode historical data of each consumer, followed by a graph convolutional layer to incorporate features in the spatial dimension. By stacking multiple layers of this unit with different parameters, the model can capture various patterns at different temporal and spatial scales. As described, it is of type \acrshort{tts}.

\paragraph{\acrfull{fc-gnn}~\cite{satorras_multivariate_2022}} This model presumes that every node is interconnected, resulting in a complete graph. However, in the \textit{Message Aggregation} step, the weight of each "message" from neighbors will be adjusted due to the attention mechanism, allowing an adaptable edge weight between nodes even though the topology of the graph does not reflect the relation between time series. In terms of forecasting, it follows the \acrshort{tts} architecture with \acrshort{mlp} as a temporal processing unit and the attention mechanism in the spatial processing unit.

\paragraph{\acrfull{bp-gnn}~\cite{satorras_multivariate_2022}} This model is a variant of the FC-GNN model. However, instead of full connectivity, it defines virtual nodes that connect to all original nodes, forming a bipartite graph. These nodes act as hubs, aggregating, updating, and relaying information between original nodes.


The selected models are arranged in Table \ref{tab:overview}, following the structure outlined in Sections \ref{subsec:graph} and \ref{subsec:processing}.

\begin{table}[h]

\caption{Summary of the models in the described framework}
\label{tab:overview}
\begingroup
\setlength{\tabcolsep}{5pt} 
\renewcommand{\arraystretch}{1.2} 
\resizebox{0.9\textwidth}{!}{
\begin{tabular}{l|ll|lll}
\hline
\multicolumn{1}{c|}{\multirow{2}{*}{Models}} & \multicolumn{2}{c|}{Graph formation}     & \multicolumn{2}{c}{Architecture} \\ \cline{2-5} 
\multicolumn{1}{c|}{}                        & Predefined graph & Learnable graph & TTS            & T\&S               \\ 
\hline
GRUGCN                                      & \multicolumn{1}{c}{\checkmark}  &                 & \multicolumn{1}{c}{\checkmark}       \\ 
\hline
GCGRU                                      & \multicolumn{1}{c}{\checkmark}  &                 &                & \multicolumn{1}{c}{\checkmark}          \\ 
\hline
T-GCN                                       & \multicolumn{1}{c}{\checkmark}  &                 &  &  \multicolumn{1}{c}{\checkmark}             \\ 
\hline
AGCRN                                      &             & \multicolumn{1}{c|}{\checkmark}      &                &  \multicolumn{1}{c}{\checkmark}            \\ 
\hline
GraphWavenet                                &     &           \multicolumn{1}{c|}{\checkmark}     & \multicolumn{1}{c}{\checkmark}               &            \\ 
\hline
FC-GNN                             &   \multicolumn{1}{c}{\checkmark}          &       & \multicolumn{1}{c}{\checkmark}    &            \\ 
\hline
Bipartite                                   &    \multicolumn{1}{c}{\checkmark}         &       & \multicolumn{1}{c}{\checkmark}    &              \\ 
\hline
\end{tabular}
}
\endgroup
\end{table}

%% file: sections/03__experiments.tex
\section{Experiments}\label{sec:exp_result}

\subsection{Dataset and data partition}
\begin{figure}[h!]
    \centering
    \includegraphics[width=0.8\linewidth]{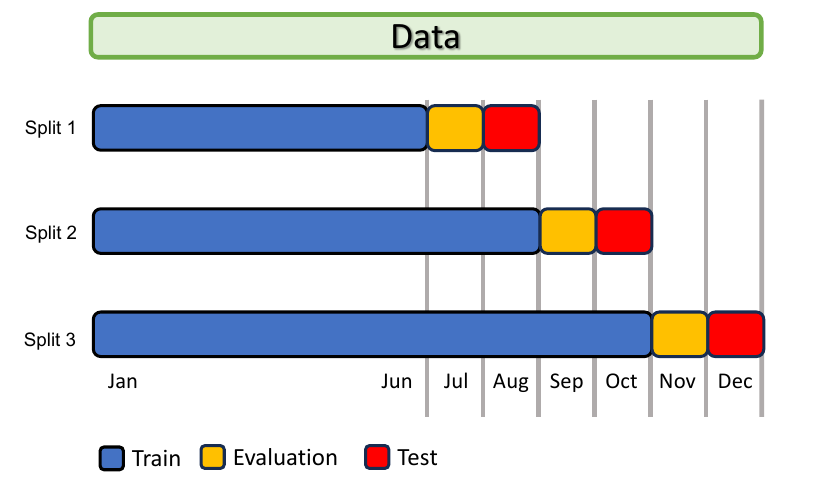}
    \caption{Train-validation-test split settings.}
    \label{fig:cross-val}
\end{figure}

 We used an open dataset to train and evaluate the performance of \acrshort{stgnn}. The dataset is \acrfull{lcl} dataset~\cite{strbac_low_2024}, containing historical smart-meter data from 5,567 households over 2013 with a 30-minute resolution. We selected 228 load profiles from one specific sociodemographic group within the dataset (Acorn - D~\cite{acorn}). We selected consumers of the same sociodemographic group so that the pairwise relationship is solely based on the historical data. The selected dataset also has no missing values and zero values, which enables the use of MAPE metrics (Section \ref{subsec:model_benchmark}). As all algorithms are designed to operate solely with past consumption data~\cite{arastehfar_short-term_2022}, we use only historical data as input for all models, as outlined in Section \ref{sec:background}. 
 For data partitioning, to avoid information leakage between training and testing, our proposed train-validation-test split is indicated in Fig. \ref{fig:cross-val}.

Specifically, the training periods for splits 1, 2, and 3 span from January 1, 2013, to the day before July 1, September 1, and November 1, respectively. The validation and testing periods of each split cover the month immediately after the training period. The partition aims to see if the performance comparison is consistent with different time scales.

\subsection{Model training}
  For all models, we do hyperparameter tuning in learning rate, batch size, training window $W$ (Section \ref{sec:background}). The other hyperparameter is fixed for each experiment. 

For training, we use MAE as the loss function.
\begin{equation}
    \text{MAE} = \sum_{n=1}^N\sum_{t=1}^W|x_i - \hat{x}_i|
\end{equation}

with $x_i$ being the measured energy consumption and $\hat{x}_i$ being the predicted consumption of consumers $i$ at time $t$. 

The maximum epoch for training is 300 with early stopping options. To facilitate the implementation of the code, we build the experiment from the tsl package~\cite{Cini_Torch_Spatiotemporal_2022}. The details of the implementation of each model can be found in specified Github repository\footnote{\textbf{\href{https://github.com/Viet1004/Benchmark_STGNN_for_STLF}{https://github.com/Viet1004/Benchmark\_STGNN\_for\_STLF}}}. All the training is carried out in the IRIS cluster of the high-performance computer (HPC) facilities of the University of Luxembourg~\cite{ORBi-aab35225-a6bc-496d-a6e7-621189ebff46}.

\subsection{Model benchmark}\label{subsec:model_benchmark}
To validate the performance of the \acrshort{stgnn} models, we compare them with several benchmark models widely used in time series forecasting.

\begin{itemize}
    \item \textbf{Seasonal Naive}: This simple baseline model uses the value of the same hour on the previous day to predict the corresponding hour on the next day. 

    \item \textbf{\acrfull{var}}: The \acrfull{var} model is a statistical approach that extends the univariate autoregression (AR) model to multivariate time series. 

    \item \textbf{\acrshort{gru}}: Recurrent Neural Networks (RNNs) are extensively used in the literature to model sequential data because of their ability to capture temporal dependencies. In this study, we use a variant of RNN, the \acrfull{gru}. This architecture is also used in many \acrshort{stgnn} models in our research.

    \item \textbf{Transformer}: Transformer models~\cite{vaswani_attention_2023} represent a paradigm shift in sequence modeling by using self-attention mechanisms to compute pairwise dependencies between elements in a sequence. This architecture has been increasingly applied to \acrshort{stlf}~\cite{zhao_spatial_2023}.
\end{itemize}

Each of these benchmark models offers unique characteristics, allowing us to comprehensively assess the performance of \acrshort{stgnn} against a diverse set of approaches that vary in complexity, interpretability, and scalability. Note that all the models above only take into account the temporal dependency in the sequence. The training configuration for these models is the same as \acrshort{stgnn} models.

The error metrics to evaluate the performance of each model are:
\begin{align}
    \text{MAE} &= \frac{1}{NT}\sum_{n=1}^N\sum_{t=1}^T|x_i - \hat{x}_i| \\
    \text{MAPE} &= \frac{1}{NT}\sum_{n=1}^N\sum_{t=1}^T \left|\frac{x_i - \hat{x}_i}{x_i}\right| \\
    \text{RMSE} &= \sqrt{\frac{1}{NT}\sum_{n=1}^N\sum_{t=1}^T\left(x_i - \hat{x}_i\right)^2}
\end{align}

where N is the number of households and T is the time steps of the testing period.

%% file: sections/04__results.tex
\section{Results and discussion}

In each experiment, we perform the procedure 5 times using identical parameters to calculate the statistics of the experiment. The results will be displayed as the mean, followed by the standard deviation from the 5 trials. To enhance the presentation of the results, we use \textbf{bold font} to highlight instances where the performance of \acrshort{stgnn} models surpasses all benchmark models, and \underline{underlining} to indicate the best performance according to the error metrics.

\subsection{Comparison between graph formation methods}\label{subsubsec:graph_formation}
Methods for graph formation can produce various topologies that facilitate the aggregation of information. This section aims to explore whether creating graphs based on different types of similarities between time series impacts forecasting performance. The focus is exclusively on the 3 first models in Table \ref{tab:overview} which require signal-based predefined graphs.
\begin{table}[h!]
\centering
\caption{Performance of models which requires predefined graph 
\newline
(Table \ref{tab:overview}) in Split 3.}
\begin{tabular}{@{}lllll@{}}
                          &     &     & Metrics  &    \\
\multirow{-2}{*}{Model}   & \multirow{-2}{*}{Graph  formation} & MAE (Wh) & MAPE (\%)    & RMSE (Wh)                                                                                        \\
\hline
                          & Euclidean        & \underline{149.7(0.1)} & 55.9(0.3) & \underline{295.6(0.6)   }\\
                          & DTW              & 149.8(0.3) & \underline{55.4(1.1)} & 295.8(1.0) \\
                          & Correntropy      & 149.8(0.4) & 55.5(0.6) & 295.7(0.7)   \\
\multirow{-4}{*}{GRUGCN}                          & Pearson          & 150.7(0.1) & 56.2(0.4) & 297.8(1.0)   \\
\hline
                          & Euclidean        & \underline{150.9(0.5)} & \underline{56.5(0.5)} & 297.9(1.3)   \\
                          & DTW              & \underline{150.9(0.4)} & 57.1(1.0) & \underline{297.3(1.3)}    \\
                          & Correntropy      & 151.4(0.6) & 56.8(0.6) & 298.8(0.7)  \\
\multirow{-4}{*}{T-GCN}                           & Pearson          & 151.7(0.3) & 57.2(0.5) & 298.6(0.4) \\
\hline
                          & Euclidean        & 149.1(0.1) & 55.7(0.4) & 295.7(0.9) \\
                          & DTW              & 149.2(0.5) & 56.2(0.6) & \underline{295.5(0.8)}     \\
                          & Correntropy      & \underline{149.0(0.4)} & \underline{55.0(0.4)} & 295.7(0.6) \\
\multirow{-4}{*}{GCGRU}                          & Pearson          & 149.6(0.2) & 56.4(0.2) & 295.8(0.6)  \\
\hline
\end{tabular}
\label{tab:graph_formation}
\end{table}

The results in \autoref{tab:graph_formation} suggest that graph construction methods do not significantly impact the effectiveness of learning, even though the resulting topologies may vary. \glspl{stgnn}, as a data-focused method, adapts by learning with different topologies to achieve comparable outcomes. In the following section, for each predefined-graph model, we employ the graph formation technique that results in the minimal MAE. 
\subsection{Model benchmark with different temporal scale at the residential level}\label{subsubsec:performance_comparison}

\begin{table*}[htbp]
\centering
\caption{Performance of different \acrshort{stgnn} models at the \textbf{residential} level. 
}
\begingroup
\setlength{\tabcolsep}{4pt} 
\renewcommand{\arraystretch}{1.2} 
\resizebox{\textwidth}{!}{
\begin{tabular}{cllllllllll}
\hline
\hline
\multirow{3}{*}{Group} & \multirow{3}{*}{Models} & \multicolumn{9}{c}{Metrics} \\ \cline{3-11}
                       &                         & \multicolumn{3}{c}{Split 1}  & \multicolumn{3}{c}{Split 2} & \multicolumn{3}{c}{Split 3} \\ \cline{3-11}
                       &                         & MAE (Wh)    & MAPE (\%)   & RMSE (Wh)   & MAE (Wh)    & MAPE (\%)    & RMSE (Wh) & MAE (Wh)    & MAPE (\%)   & RMSE (Wh) \\ 
\hline
\multirow{4}{*}{Benchmark} 
& SeasonalNaive                              & 114.4(0.0) & 66.4(0.0) & 238.6(0.0) & 156.4(0.0) & 75.9(0.0) & 304.1(0.0) & 186.3(0.0) & 81.1(0.0) & 358.2(0.0) \\
& VAR                                         & 134.3(0.2) & 109.5(0.5) & 218.5(0.3) & 167.6(0.1) & 106.7(0.2) & 283.9(0.4) & 210.1(0.5) & 134.7(0.5) & 335.0(0.6)      \\
& GRU &
89.5(0.1) & 44.7(0.3) & 194.0(0.5) & 126.5(0.2) & 50.5(0.4) & 254.7(0.4) & 153.1(1.7) & 56.9(2.2) & 302.2(0.9) \\
& Transformer                                  &    90.2(0.3) & 44.9(0.8) & 194.5(0.4) & 127.9(0.5) & 52.1(0.4) & 255.4(0.5) & 154.1(0.1) & 56.7(0.1) & 303.4(0.4)      \\
\hline
\multirow{8}{*}{STGNN}  & GRUGCN                                         & \textbf{89.0(0.2)} & \underline{\textbf{43.0(1.1)}} & \textbf{193.7(0.6)}	     & \textbf{125.3(0.2)} & \textbf{50.0(0.2)} & \textbf{251.4(0.5)}       & \textbf{149.7(0.1)} & \textbf{55.9(0.3)} & \textbf{295.6(0.6)   }    \\
& GCGRU                                      & \underline{\textbf{88.2(0.2)}} & \textbf{43.6(0.4)} & \textbf{191.9(0.2)}     & \textbf{124.6(0.2)} & \underline{\textbf{49.3(0.6)}} & \textbf{251.9(0.4)}   &    \textbf{149.0(0.4)} & \textbf{55.0(0.4)} & \textbf{295.7(0.6)}  \\
& T-GCN                                     & \textbf{88.9(0.5)} & \textbf{43.8(0.1)} & \textbf{193.5(1.0)}   &   \textbf{125.0(0.5)}      &   \textbf{49.7(0.6)}      &     \textbf{253.7(1.1)}    & \textbf{150.9(0.5)} & \textbf{56.5(0.5)} & \textbf{297.9(1.3)}  \\
& AGCRN                                     & 91.0(0.2) & 49.1(0.4) & \textbf{190.9(0.6)} & \textbf{123.8(0.3)} & 50.6(0.6) & \textbf{248.7(0.6)} & \textbf{150.1(0.3)} & 58.8(0.2) & \textbf{294.2(1.1)}       \\
& GraphWavenet   & 91.4(0.2) & 48.1(0.8) & 196.7(0.8) & 128.4(0.3) & 52.9(0.6) & 256.7(1.0) & 155.9(0.5) & 60.1(0.5) & 304.6(0.8)  \\
& FC-GNN                            &   \textbf{88.6(0.1)} & 45.7(0.6) & \textbf{189.0(0.3)} & \textbf{121.7(0.4)} & \textbf{49.5(0.6)} & \underline{\textbf{245.9(1.5)}} & \underline{\textbf{146.9(0.1)}} & \textbf{55.6(0.6)} & \underline{\textbf{290.3(1.0)}}     \\
& BP-GNN                                   &   90.2(0.1) & 48.8(0.5) & \underline{\textbf{188.4(0.3)}} & \underline{\textbf{121.6(0.1)}} & \textbf{49.4(0.5)} & \textbf{246.0(0.8)} & \textbf{148.0(0.1)} & \underline{\textbf{54.7(0.5)}} & \textbf{293.3(0.9) }     \\
\hline
\hline
\end{tabular}
}
\endgroup
\label{tab:result_228}
\end{table*}

Compared to benchmark models that process only temporal features, the practice of adding relationships between households in \acrshort{stgnn} models increases forecasting performance. Especially, in contrast to \acrshort{gru}, which uses matrix multiplication as the unit cell of the recurrent network, \acrshort{gcgru} and \acrshort{t-gcn} always achieve better performance by applying \acrshort{gcn} as the unit cell. Similarly, \acrshort{grugcn} also outperforms \acrshort{gru} by applying \acrshort{gcn} on top of it to account for spatial dependency. It showcases the effective use of graph neural networks to model spatial relationships (see Table \ref{tab:result_228}).


Moreover, although models with a learnable graph theoretically offer more flexibility, this approach does not deliver promising outcomes; only \acrshort{agcrn} model performs compatible results with the best benchmark models (\acrshort{gru}). However, when comparing with \acrshort{gcgru}, which uses a predefined graph from signals, there is a downgrade in performance in some metrics. This is because the learnable graph offers a more flexible way to model the spatial relationship. However, it could make the forecasting task more susceptible to overfitting. When testing on a more distant future (one month after training, as outlined in Fig. \ref{fig:cross-val}), the model actually performs worse than the counterpart that uses a predefined graph. 

When testing with 3 splits, the \acrshort{bp-gnn} and \acrshort{fc-gnn} models often perform better than their counterparts. These models presume the topology of the graph without relying on the data (fully connected or bipartite); instead, when aggregating information from neighbor nodes, the models utilize a weighted sum of neighboring information, where the weights are derived from the input. This makes the "message aggregation" step more adaptable to the input. 
However, despite the naive assumption of graph topology, their better performance questions if the graph formation based on signals or learnable parameters is effective in the context of \acrshort{stlf}. 
An interesting case is the \acrshort{bp-gnn} model. Although it performs slightly worse than \acrshort{fc-gnn}, it is more scalable since the interaction between the nodes of \acrshort{fc-gnn} is $N^2$, while for the \acrshort{bp-gnn} model, it is only $2KN$ with $K$ being the number of virtual nodes. It also outperforms other models most of the time. One reason might be that, due to the nature of the load profile dataset, the consumption pattern of users is grouped by latent factors such as socio-demographic status~\cite{acorn}. The virtual nodes defined by the model can account for latent factors that can influence the original nodes (households). The bipartite topology allows these virtual nodes to gather information in a cluster-like manner; then, the aggregated information is passed down to each node as additional information for learning.


\subsection{Model benchmark with different temporal scale at the aggregate level}
\begin{figure}[h!]
    \centering
    \includegraphics[width=0.6\linewidth]{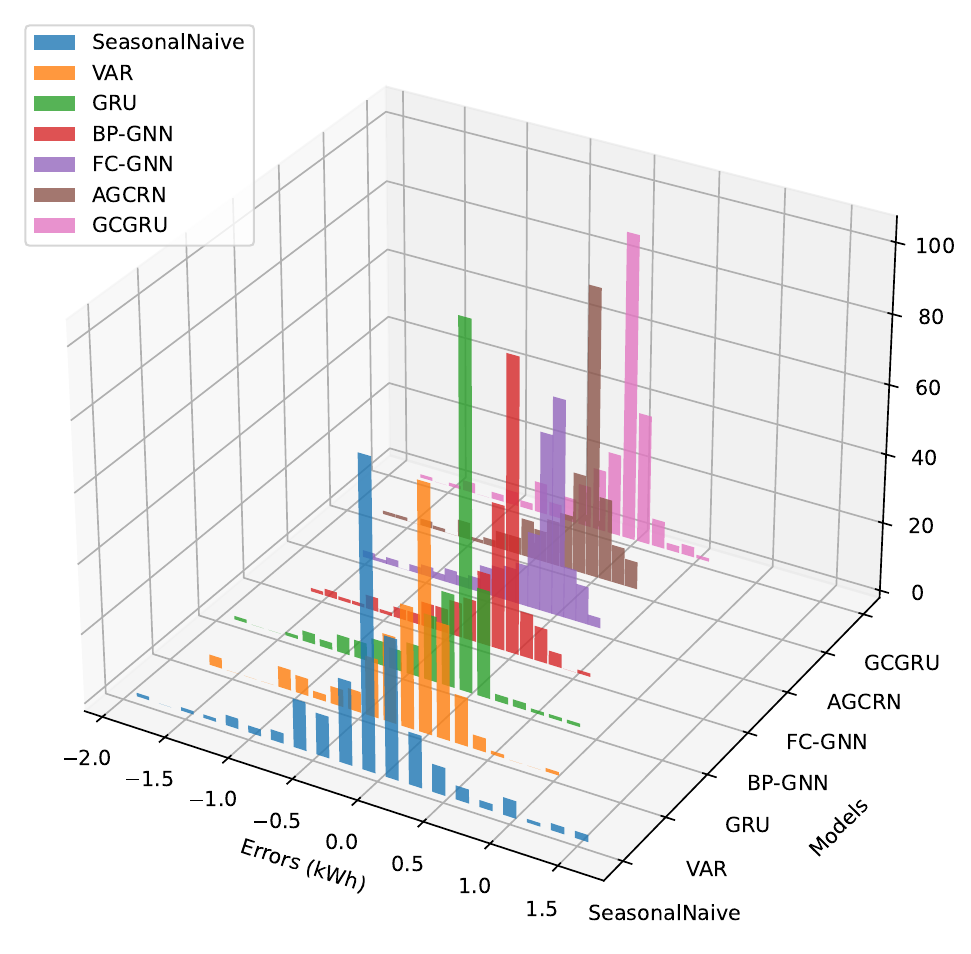}
    \caption{Distribution of differences between forecasts and ground truth.}
    \label{fig:multiple-forecast}
\end{figure}

We investigate the performance of load forecasting at the aggregate level simply by aggregating all the forecasts at the residential level (see Table \ref{tab:result_agg_228}). At the aggregate level, the forecast results obtained through aggregation are inferior to those of the baseline model (SeasonalNaive). This behavior is also observed in other deep learning models such as \acrshort{gru} and \acrshort{tfm}. 
To explain this behavior, we visualize in Fig.\ref{fig:multiple-forecast} the histogram of the errors among all the households at peak hour (2013-12-23 19:00). The x-axis represents errors, the y-axis lists top-performing models (Table \ref{tab:result_agg_228}), and the z-axis shows frequency.

\begin{table*}[htbp]
\centering
\caption{Performance of different \acrshort{stgnn} models at the \textbf{aggregate} level.
}
\begingroup
\setlength{\tabcolsep}{3pt} 
\renewcommand{\arraystretch}{1.2} 
\resizebox{\textwidth}{!}{
\begin{tabular}{cllllllllll}
\hline
\hline
\multirow{3}{*}{Group} & \multirow{3}{*}{Models} & \multicolumn{9}{c}{Metrics} \\ \cline{3-11}
                       &                         & \multicolumn{3}{c}{Split 1}  & \multicolumn{3}{c}{Split 2} & \multicolumn{3}{c}{Split 3} \\ \cline{3-11}
                       &                         & MAE (kWh)    & MAPE (\%)   & RMSE (kWh)   & MAE (kWh)    & MAPE (\%)    & RMSE (kWh) & MAE (kWh)    & MAPE (\%)   & RMSE (kWh) \\ 
\hline
\multirow{4}{*}{Benchmark} 
& SeasonalNaive                             & \underline{3.39(0.0)} & \underline{6.8(0.0)} & \underline{4.57(0.0)} & \underline{5.48(0.0)} & \underline{8.2(0.0)} & \underline{7.96(0.0)} & \underline{5.97(0.0)} & \underline{7.6(0.0)} & \underline{8.47(0.0)} \\
& VAR                                         & 3.44(0.23) & 7.1(0.05) & 4.56(0.04) & 13.73(0.25) & 18.0(0.4) & 18.67(0.27) & 9.56(0.19) & 11.5(0.2) & 12.84(0.24) \\
& GRU  & 8.58(0.28) & 16.9(0.6) & 10.08(0.29) & 13.49(0.17) & 18.8(0.3) & 16.97(0.17) & 15.35(0.39) & 18.2(0.7) & 18.76(0.29) \\
& Transformer                                 & 8.83(0.37) & 17.3(0.6) & 10.34(0.43) & 13.34(0.73) & 18.6(0.2) & 16.91(0.05) & 15.72(0.17) & 18.5(0.2) & 19.30(0.12) \\
\hline
\multirow{8}{*}{STGNN} 
& GRUGCN                                      & 9.38(0.57) & 18.2(1.2) & 10.93(0.58) & 12.71(0.31) & 17.7(0.4) & 16.17(0.4) & 14.31(0.37) & 16.7(0.3) & 17.71(0.44) \\
& GCGRU                                       & 8.88(0.54) & 17.1(1.1) & 10.34(0.51) & 13.17(0.54) & 18.2(0.90) & 16.87(0.53) & 14.34(0.17) & 16.9(0.3) & 17.74(0.20) \\
& T-GCN                                       & 9.27(0.26) & 18.1(0.5) & 10.68(0.28) & 13.93(0.33) & 19.4(0.4) & 17.58(0.41) & 14.52(0.44) & 17.4(0.4) & 17.81(0.55) \\
& AGCRN                                     & 7.91(0.15) & 15.2(0.3) & 9.43(0.19) & 13.26(0.35) & 17.9(0.5) & 17.34(0.34) & 13.16(0.28) & 15.3(0.2) & 16.66(0.36) \\
& GraphWavenet   & 8.85(0.44) & 17.4(1.0) & 10.34(0.44) & 13.91(0.47) & 19.2(0.9) & 17.61(0.43) & 15.02(0.44) & 18.1(0.5) & 18.37(0.57) \\   
& FC-GNN                             & 8.22(0.32) & 15.9(0.7) & 9.80(0.27) & 12.79(0.56) & 16.9(0.7) & 16.79(0.72) & 13.61(0.2) & 16.0(0.4) & 16.60(0.14) \\
& BP-GNN                                   & 6.61(0.35) & 13.1(0.8) & 8.07(0.34) & 12.77(0.39) & 17.1(0.6) & 16.56(0.42) & 14.81(0.29) & 16.9(0.5) & 18.40(0.24) \\
\hline
\hline
\end{tabular}
}
\endgroup
\label{tab:result_agg_228}
\end{table*}

We observe that the baseline model, although predicting less accurately (errors are more scattered), has its tails more evenly distributed. The error of deep learning-based models (including \glspl{stgnn}) is often more skewed on the left. An explanation is that these models tend to prioritize learning from "easier" periods with consistent patterns while failing to adequately capture atypical events, such as consumption spikes~\cite{zhang_unlocking_2023} (see Fig. \ref{fig:multiple-forecast}). This can lead to underestimation during abnormal periods. At the aggregate level, instead of centering predictions around the actual average consumption, the forecasts may consistently underestimate the consumption spikes, and hence the errors will not be balanced out. In \glspl{stgnn} since it incorporates spatial learning, it can potentially propagate errors throughout the spatial dimension and therefore does not address this issue.



%% file: sections/05__conclusions.tex
\section{Conclusion}\label{sec:conclusion}

In this paper, we provide an overview of the literature on \glspl{stgnn} in short-term load forecasting and benchmark selected algorithms. Our findings show that integrating spatial relationships with temporal features improves forecasting accuracy for household energy consumption compared to models using only temporal features. Nevertheless, the most effective method for creating a graph to represent the proximity of energy consumption among households remains undetermined, given that even rudimentary graphs, like bipartite or fully connected graphs, can perform better than signal-based graphs. We also discuss how different components of \glspl{stgnn} can affect the performance of models. Notably, bipartite graphs effectively capture energy consumption dynamics, outperforming direct relationships from raw signals or embeddings. However, at the aggregate level, simple models such as SeasonalNaive outperform \gls{stgnn} models.


However, our research has its limitations; we acknowledge that, for instance, our forecasting scenarios focus on one forecasting horizon (day ahead), which may not provide a comprehensive comparison of the strengths and weaknesses of the models. Furthermore, we did not consider the incorporation of exogenous variables into the STGNN models. Since energy consumption is significantly influenced by exogenous factors, such as weather or time indicators, the inclusion of these variables could improve model performance and provide deeper insight. Finally, given the growing scientific literature around \acrshort{stgnn} for residential \acrshort{stlf}, our selected models for benchmarking may not represent the entire research landscape in this domain. We believe that this incompleteness can motivate future research to provide a broader overview of \acrshort{stgnn} into the residential \acrshort{stlf} problem.